# A Novel Robust Control Method Combining DNN-Based NMPC Approximation and PI Control: Application to Exoskeleton Squat Movements

**Alireza Aliyari\*, and Gholamreza Vossoughi**

**Abstract:** Nonlinear Model Predictive Control (NMPC) is a precise controller, but its heavy computational load often prevents application in robotic systems. Some studies have attempted to approximate NMPC using deep neural networks (NMPC-DNN). However, in the presence of unexpected disturbances or when operating conditions differ from training data, this approach lacks robustness, leading to large tracking errors. To address this issue, for the first time, the NMPC-DNN output is combined with a PI controller (Hybrid NMPC-DNN-PI). The proposed controller is validated by applying it to an exoskeleton robot during squat movement, which has a complex dynamic model and has received limited attention regarding robust nonlinear control design. A human-robot dynamic model with three active joints (ankle, knee, hip) is developed, and more than $5.3 \times 10^6$ training samples are used to train the DNN. The results show that, under unseen conditions for the DNN, the tracking error in Hybrid NMPC-DNN-PI is significantly lower compared to NMPC-DNN. Moreover, human joint torques are greatly reduced with the use of the exoskeleton, with RMS values for the studied case reduced by 30.9% at the ankle, 41.8% at the knee, and 29.7% at the hip. In addition, the computational cost of Hybrid NMPC-DNN-PI is 99.93% lower than that of NMPC.

**Keywords:** Nonlinear Model Predictive Control (NMPC), Deep Neural Network (DNN), Hybrid Control, Exoskeleton Robot, Human-Robot Interaction.

## 1. INTRODUCTION

Model predictive control (MPC) is an advanced control method that has gained a prominent position in industry [1]. Its main advantages include a forward-looking approach, better performance compared to many other controllers, easy scalability to multi-input multi-output systems, and applicability to systems with delays and non-minimum phase behavior [2–6]. Nonlinear model predictive control (NMPC) provides a powerful approach for controlling nonlinear dynamic systems. In this type of control, the control signal is calculated either by online linearization combined with linear MPC or through numerical optimization of the nonlinear system. When online linearization cannot accurately represent the behavior of the nonlinear system, the optimization problem is solved directly for the nonlinear system. In this case, unlike linear systems, the optimal control problem is not necessarily convex, and analytical solutions are often not available [7]. Therefore, numerical optimization is required to minimize the cost functions.

Many studies have explored the use of NMPC in various dynamic systems, demonstrating the remarkable capability of this control method in controlling different types of systems [8-13]. The main limitation of NMPC is its high computational time, which often makes it practically impossible to implement on current hardware for controlling robotic systems. To address this issue, numerous studies have investigated approximating the behavior of model predictive controllers using deep neural networks for different dynamic systems. In earlier years, attempts to approximate MPC using neural networks failed due to shallow network architectures and the lack of powerful processors. However, in recent years, with the advent of deep neural networks and advances in computing hardware, researchers have revisited this approach, and it remains an open research topic in the field.

Moriyasu et al. [11] applied a nonlinear model predictive control system to regulate airflow in diesel engines. They solved the optimization problem using the Sequential Quadratic Programming (SQP) numerical optimization method and reduced the computational load by approximating the control system with a deep neural network. Lucia et al. [14] also used a deep neural network to approximate the behavior of a multi-stage model predictive control system for controlling a chemical

Alireza Aliyari and Gholamreza Vossoughi are with the School of Mechanical Engineering, Sharif University of Technology, Tehran, Iran (Emails: alireza.aliyari@alum.sharif.edu, vossough@sharif.edu).
\* Corresponding author.

process. However, the amount of data provided to the neural network in their study was very limited, making it impractical in real applications, since even small deviations from the training data could lead to system instability. Tagliabue et al. [15] carried out a similar study for controlling a robotic system and significantly reduced the time required to compute the control signal. A major challenge with these studies, and with neural-network-based NMPC approximation in general, is that although the original NMPC may be robust, the approximated controller is not necessarily robust. In the presence of disturbances or deviations from training data, the system may become unstable. Therefore, the first objective of the present study is to address this issue by combining the approximated NMPC with a PI controller, providing a significant step toward implementing predictive control systems with very low computational cost on modern hardware. From here onward, this controller will be referred to as Hybrid NMPC-DNN-PI.

In recent years, research interest in robotic systems has increased across diverse fields of control and automation [16-18], highlighting the growing demand for reliable and efficient control strategies. Within this context, the proposed control system is validated on an exoskeleton robot to reduce human joint torques during squatting. Squatting is a physically demanding activity that puts significant stress on human joints. Performing this movement for long periods leads to fatigue and reduces the ability to continue tasks, while also increasing the risk of musculoskeletal disorders. Using lower-limb wearable robots during squatting can help prevent fatigue and reduce such problems [19]. Wearable robots interact directly with humans, so they need an accurate control system to ensure safety. A precise and intelligent control design enables the robot to support humans in heavy tasks and reduce the pressure on the joints, while an inaccurate design can place extra pressure on them. These robots should be controlled in a way that does not interfere with human movement and helps reduce the amount of force required from the muscles. The dynamic equations governing squatting are nonlinear, requiring the design and implementation of complex nonlinear control systems. Moreover, due to the high torques needed from the motors during squatting, errors in calculating the control signal can cause serious injuries to users. Moreover, due to the need for rapid calculation of the control signal for controlling the exoskeleton robots, this case provides a very suitable example for implementing the proposed control system, as it strongly requires such an accurate and fast controller.

Despite the importance of designing an appropriate control system for lower-limb wearable robots specifically for squatting, studies on this topic are limited. Wei et al. [20] studied the control of the hip joint in a lower-limb wearable robot during semi-deep squatting. They used a linear PD for controlling the robot. Sado et al. [21] employed a linear LQG control system to control an exoskeleton with two active joints during squatting. They obtained the required human joint torques using the inverse dynamics method based on a human motion model. However, the control systems applied in these studies were linear and not robust against the uncertainties present in the system. Luo et al. [22] used a reinforcement learning-based controller to control the robot during squatting and to assist in maintaining human balance. Their control system was only suitable for rehabilitation purposes and could not be generalized to power-augmentation situations, since they provided the controller with pre-defined reference signals. This approach cannot be applied when the human instantly decides to take different movement paths. Therefor, there is no precise nonlinear control system in the literature for controlling an exoskeleton robot during squatting for power-augmentation purposes, without the need to predefine reference motion signals, and in which the robot acts based on human movement. Consequently, the second objective of this study is to fill this gap. This also serves as a validation of the proposed control system.

The rest of this paper is organized as follows: Section 2 presents a dynamic model for applying the control system, which includes both the robot and human models during squat motion. Section 3 discusses the proposed control system, and the results are presented in Section 4. Finally, Section 5 provides the conclusion of the study.

## 2. DYNAMIC MODEL

In this section, the dynamic model of the robot and human during squatting is examined as a study subject with complex dynamics to enable the implementation of the controller. The presented calculations and equations can be generalized and applied to other dynamic systems as well. The goal of the control system is to calculate the control signals and assist the human during squatting in the sagittal plane, where most of the human joint torques occur. Therefore, the motion equations of both the human and the robot are analyzed within this plane.

The dynamic models of the robot and the human in the sagittal plane are similar and consist of three active joints: the ankle, knee, and hip, namely the shank, thigh, and upper body. During squatting, the foot remains fixed on the ground, and the position of the ankle joint does not change. Fig. 1 shows the schematic of the joints and links, along with the dynamic parameters. The angles of the ankle, knee, and hip joints are denoted as $\theta_1$, $\theta_2$, and $\theta_3$, respectively. Meanwhile, the lengths of the robot's links, the distances to their centers of mass, the masses, and the inertias of the links are denoted by $L_j$, $L_{cj}$, $m_j$, and $I_j$, respectively (j=1, 2, 3).

The dynamic equation of the robot during squatting, derived from the Euler-Lagrange equations, is presented in Eq. 1.

$$M_R(\theta_R)\ddot{\theta}_R + C_R(\theta_R, \dot{\theta}_R)\dot{\theta}_R + G_R(\theta_R) + d = u. \qquad (1)$$

Where $\theta_R = [\theta_{1R} \ \theta_{2R} \ \theta_{3R}]$ represents the robot joints angle vector, $M_R$ the inertia matrix, $C_R$ the Coriolis matrix, and $G_R$ the gravity matrix. The total torque applied to the robot joints is expressed as $u = T_R + T_{int}$, where $T_R$ is the torque generated by the robot actuators, and $T_{int}$ is the torque resulting from human-robot interaction forces, transmitted through the straps and measurable by force sensors. Furthermore, $d$ denotes the sum of all unmodeled terms, uncertainties and noises in the system. For the human model, the total torque is given as $T_R - T_{int}$, and the other parts of Eq. 1 are similar. The straps installed between the human and the robot are responsible for transmitting forces.

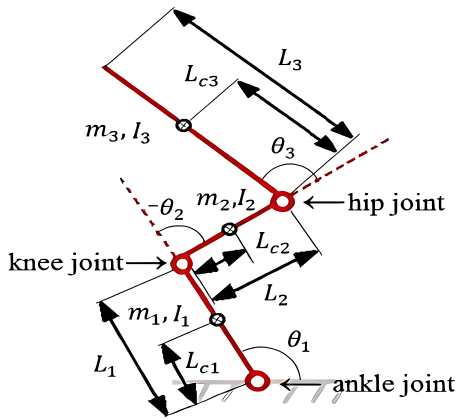

Fig. 1. Dynamic model of human/wearable robot.

## 3. CONTROL DESIGN

In this section, the designed control system is examined. First, the nonlinear model predictive control (NMPC) is explained, and then the proposed control system (Hybrid NMPC-DNN-PI) is discussed.

### 3.1. Nonlinear model predictive control

The model predictive control system calculates the control signal by considering the constraints of the problem, using the current and past values of the state variables, as well as their desired values for the upcoming steps, through the dynamic model of the system. This control method predicts the future state of the system and computes the control signals accordingly. To implement discrete MPC, the robot's continuous-time model is converted into a discrete-time model using the fourth-order Runge-Kutta method. Fig. 2 illustrates the structure of the model predictive control system. It is assumed that the system is at the k-th instant, where the state variables are represented by $x(k)$. As shown in this figure, the state variables at time $k$ and the previous steps are available, and their values need to be predicted for $P$ future steps, which is referred to as the prediction horizon. This prediction is made based on the control signals $u(k)$ to $u(k + M)$ and the system dynamics. Here, $M$ represents the control horizon, which is smaller than or equal to the prediction horizon $P$.

At each time step, the control signals $u(k)$ to $u(k + M)$ are calculated in such a way that the cost function defined in the MPC is minimized. Then, only the first control signal $u(k)$ is applied to the system, and the same procedure is repeated in the following time steps.

The accuracy of the optimal control signal found by the optimizer, the speed of convergence to the global minimum point, and avoiding convergence to local minima play an important role in choosing a suitable optimization method for the cost function in nonlinear predictive control. Traditional numerical methods, such as Newton's method may lead the cost function to converge to local minima, causing errors in the system and leading to instability. To address this issue, more advanced methods suitable for non-convex optimization can be used.

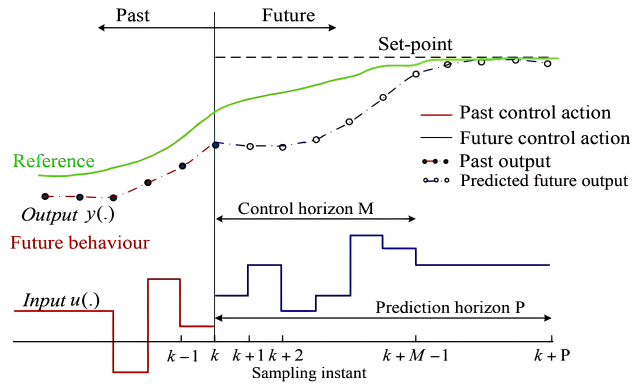

Fig. 2. The structure of the model predictive control system [23].

The Sequential quadratic programming (SQP) method [24] is one such approach, which solves non-convex optimization problems by iteratively solving Quadratic Programming (QP) subproblems. This optimization method is also effective in handling constraints. In the case of exoskeleton robot control, the applied constraints are used to limit the rate of change of torque in order to achieve a smooth control system.

### 3.2. Control objective

The cost function in the model predictive control system represents the control objective and is formulated as shown in Eq. 2.

$$J = (y_d - y)^T R_1 (y_d - y) + \Delta u^T R_2 \Delta u. \tag{2}$$

Where:

$$y = [\, \theta_{1R}(k + 1|k), \theta_{2R}(k + 1|k), \, \theta_{3R}(k + 1|k), \\ \dot{\theta}_{1R}(k + 1|k), \, \dot{\theta}_{2R}(k + 1|k), \, \dot{\theta}_{3R}(k + 1|k), \, \dots, \\ \theta_{1R}(k + P|k), \, \theta_{2R}(k + P|k), \, \theta_{3R}(k + P|k), \\ \dot{\theta}_{1R}(k + P|k), \dot{\theta}_{2R}(k + P|k), \, \dot{\theta}_{3R}(k + P|k)\,]. \tag{3}$$

Furthermore, $y_d$ represents the desired trajectory of the output vector $y$ and $\Delta u$ denotes the control signal increment over the control horizon, while $R_1$ and $R_2$ are constant matrices used to adjust the influence of different terms in the cost function.

The desired angles and angular velocities of the robot should be set in such a way that the robot assists the human during squatting and reduces the torque generated in the joints by the muscles. The total torque in human joints is the result of gravitational, Coriolis, and inertial torques. During squatting, the dominant effect comes from gravitational torques. Unlike inertial and Coriolis torques, gravitational torques do not depend on joint speed or acceleration but only on joint angles. This makes gravitational torques smoother than the other terms and prevents them from showing sudden jumps in special situations, such as when the human changes direction abruptly. The control system is designed to generate desired interaction torques that cancel out the human's gravitational torques. Since the desired interaction torques are smooth, the robot's control system avoids dangerous and harmful jumps, a situation that may occur if the system attempted to cancel the Coriolis and inertial terms.

First, the human gravitational torques need to be estimated. To maintain user comfort, no encoders are installed on the human legs to directly measure the exact joint angles, which are needed for calculating these torques. Instead, the joint angles of the robot are used to estimate them. Although this introduces some error in calculating the exact value of the human gravitational torques, the difference between the actual human torques and the estimated ones has only a limited effect on the control system of the robot. The desired interaction torques, which are considered equal to the negative of the estimated gravitational torques, are given in Eqs. 4 to 7.

$$T_{int1d} = -G_{1e} = -[m_{2h}gL_{c2h}\cos(\theta_1 + \theta_2) + m_{1h}gL_{c1h}\cos(\theta_1) + (m_{2h} + m_{3h})gL_{1h}\cos(\theta_1) + m_{3h}gL_{c3h}\cos(\theta_1 + \theta_2 + \theta_3) + m_{3h}gL_{2h}\cos(\theta_1 + \theta_2)]. \tag{4}$$

$$T_{int2d} = -G_{2e} = -[m_{2h}gL_{c2h}\cos(\theta_1 + \theta_2) + m_{3h}gL_{c3h}\cos(\theta_1 + \theta_2 + \theta_3) + m_{3h}gL_{2h}\cos(\theta_1 + \theta_2)]. \tag{5}$$

$$T_{int3d} = -G_{3e} = -m_{3h}gL_{c3h}\cos(\theta_1 + \theta_2 + \theta_3). \tag{6}$$

$$T_{intd} = \begin{bmatrix} T_{int1d} \\ T_{int2d} \\ T_{int3d} \end{bmatrix}. \tag{7}$$

Here, $T_{intd}$ is the desired interaction torque, $G_{1e}$, $G_{2e}$, and $G_{3e}$ are the estimated human gravitational torques at the ankle, knee, and hip joints, respectively. $L_{jh}$, $L_{cjh}$ and $m_{jh}$ are the lengths of the human links, the distances to their centers of mass, and the masses of these links, which are calculated based on the human's total weight and height according to [25]. The desired interaction torques can easily be converted into desired interaction forces using the torque arm. With these forces, the desired angles and velocities that are introduced to the robot's control system can be calculated using Eqs. 8 and 9.

$$\theta_d = \theta + C_1(F_{int} - F_{intd}). \tag{8}$$

$$\dot{\theta}_d = \dot{\theta} + C_2(F_{int} - F_{intd}). \tag{9}$$

Here, $\theta_d$ and $\dot{\theta}_d$ are the desired vectors of joint angular positions and velocities, respectively. $F_{int}$ and $F_{intd}$ denote the actual and desired human-robot interaction forces, respectively. Moreover, $C_1$ and $C_2$ are defined as diagonal matrices with non-negative entries.

### 3.3. Hybrid NMPC-DNN-PI control framework

Since deep neural networks can approximate complex functions [26], they can be used to model the behavior of the control systems. This approach significantly reduces the heavy computational load of nonlinear model predictive control, and once the network is trained, the time required to compute the control signal becomes negligible. The neural network establishes a relationship between the inputs and outputs of the control system. In this way, by providing the joint angles and velocities, and actual and desired interaction torques as inputs to the neural network, the required joint torques can be obtained as its outputs.

In this study, a feedforward deep neural network with three hidden layers and 50 neurons per layer is used to approximate the behavior of the NMPC. The network is trained using the Resilient Backpropagation method, and an adaptive factor is applied to adjust the learning rate during training to increase the convergence speed [27]. A bipolar sigmoid function is used as the activation function for the neurons. The sigmoid function is the most commonly used activation function in neural networks

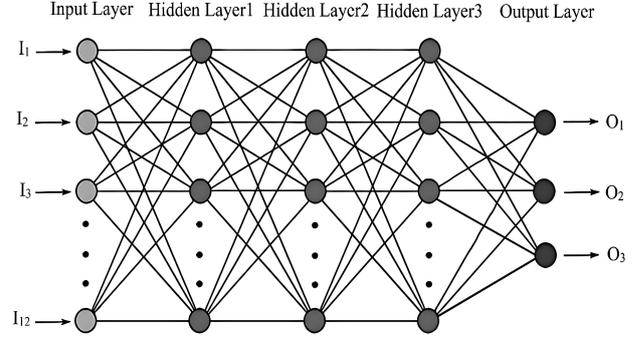

**Fig. 3.** The proposed feedforward deep neural network.

[28]. It has a smooth derivative and is well-suited for networks with a moderate number of layers. To improve the network's accuracy in learning the relationships between the data, the input data are scaled to the range [-1, 1]. The neural network used in this study is shown in Fig. 3.

To train the deep neural network, simulations were performed on the squatting motion models of both humans and the robot. For this purpose, 25 trajectories from the dataset [29] were introduced to the simulated human model. The human model attempts to follow these signals using a feedback-linearization controller, which is employed to model the function of the human brain. In this way, a dynamic human model is created, whose motion and the torques generated in its joints are also influenced by the robot's movement and the interaction forces between the human and the robot. The robot control system, without prior knowledge of these trajectories, can control the robot and reduce the torques generated in the human joints solely through the interaction forces. Additionally, nine different combinations of human weight and height were considered, each tested at six different squatting speeds (1.5, 2, 2.5, 3, 4, and 5 seconds for a full squatting cycle), as presented in Table 1.

Data collection for the control system was carried out at different control time steps so that the network could be trained on a wide range of data, including higher time steps where the tracking error of the desired output is larger. This helps improve the robustness of the DNN-based control system. In total, more than $5.3 \times 10^6$ data samples per input were provided to the neural network for training. Fig. 4 shows the changes in Mean Squared Error (MSE) on the training data over different training epochs. The final MSE reached about 0.070% after 275 training epochs.

Despite training the network on data with different tracking errors, there is still a risk of instability because the control system itself is not robust. At the beginning, when the network inputs match the training data, the network performs well in predicting control signals. However, as time progresses, due to unexpected disturbances and accumulated errors, the similarity between the current inputs and the training data gradually decreases, which can lead to increased control error and even instability. This issue is particularly dangerous for an exoskeleton robot that directly interacts with a human. To solve this problem, the output of a PI controller is added to the value predicted by the network. The PI controller serves as the fine stage of the control system and system stabilizer, while the DNN-based approximated NMPC provides the coarse regulation. The total torque calculated

Table 1. Diffrent combinations of human weight and height used for simulation.

| Human weight / height | | |
|---|---|---|
| 60 kg / 1.55 m | 60 kg / 1.65 m | 60 kg / 1.70 m |
| 75 kg / 1.60 m | 75 kg / 1.70 m | 75 kg / 1.75 m |
| 90 kg / 1.65 m | 90 kg / 1.75 m | 90 kg / 1.80 m |

by the Hybrid NMPC-DNN-PI control system is given in Eq. 10.

$$T_R(k) = O(k) + K_P\big(F_{int}(k) - F_{intd}(k)\big) + K_I \sum_{i=1}^{k}\big(F_{int}(i) - F_{intd}(i)\big). \quad (10)$$

Where $T_R = [T_{R,ankle}\ T_{R,knee}\ T_{R,hip}]$ is the total applied torque vector, $O = [O_1\ O_2\ O_3]$ is the torque vector predicted by the deep neural network, and $K_P$ and $K_I$ are the proportional and integral gains of the PI controller.

## 4. RESULTS

In this section, to validate the proposed control system (Hybrid NMPC-DNN-PI), it is applied to the exoskeleton robot, and its performance is compared with that of the NMPC approximated by a deep neural network (NMPC-DNN) under the same conditions. To test the robustness of the controller against disturbances, the value of $d$ in Eq. 1 is set to $5\sin(t) + 0.2\sin(1000t + \frac{\pi}{2})$. The human weight and height are assumed to be 80 kg and 1.9 m, and the squat cycle is set to 1.75 s, so the controller is tested under conditions completely different from the training data. The prediction horizon and control horizon in the NMPC are both set to 3, with a maximum control signal change of 5 Nm per time step. The control signal is applied with a time step of 2 ms. The constants $K_P$ and $K_I$ in Eq. 10 are set to 0.2 and 0.13, respectively. The matrices $R_1$ and $R_2$ in Eq. 2 are defined as $\big(I_3 \otimes diag(1,1,1,0.05,0.05,0.05)\big)$ and $10^{-5}I_3$, respectively. Finally, the matrices $C_1$ and $C_2$ in Eqs. 8 and 9 are chosen as $\begin{bmatrix} 0.05 & 0 & 0 \\ 0 & 0.05 & 0 \\ 0 & 0 & 0.1 \end{bmatrix}$ and $\begin{bmatrix} 0.01 & 0 & 0 \\ 0 & 0.01 & 0 \\ 0 & 0 & 0.03 \end{bmatrix}$, respectively.

The obtained results are presented in Figs. 5 to 8. Fig. 5 shows the tracking of the desired interaction torque by the Hybrid NMPC-DNN-PI and NMPC-DNN control systems for the shank, thigh, and upper body. As can be seen, despite the applied uncertainties and unseen conditions for the deep neural network, the proposed controller, acting as a robust system, successfully followed the desired signals and recorded only small tracking errors. In contrast, the NMPC-DNN controller, due to its lack of robustness, failed to track the desired signals. These results highlight the clear advantage of the proposed controller compared with common approaches for approximating NMPC controllers while reducing computational cost.

Fig. 6 shows the torque calculated by the proposed control system, along with the contributions from the DNN and the added PI term for the robot joints. According to these results, the majority of the control signal is generated by the deep neural network, while the PI term plays a smaller role, compensating for errors from the network to achieve a stable and robust control system. The angular positions and velocities of the robot's ankle, knee, and hip joints are shown in Fig. 7. The displayed velocities vary smoothly, indicating the smooth behavior of the control system.

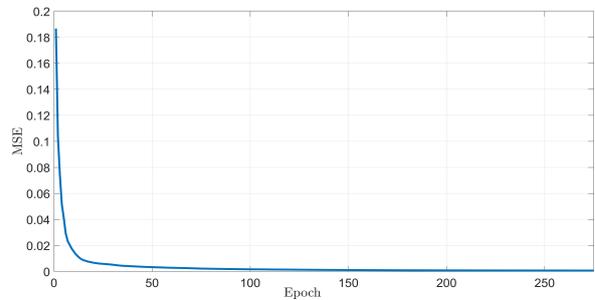

Fig. 4. Changes in MSE on the training data over different training epochs.

To evaluate the effect of the Hybrid NMPC-DNN-PI controlled exoskeleton on human effort during the squat movement, the torques generated in the human joints for two cases: without the robot and with the robot, are presented in Fig. 8. According to these results, using the exoskeleton, whose controller was designed to compensate for human gravitational torques, effectively assisted the human and reduced joint torques. For the human case studied, and a squat period of 1.75 s, the RMS values of the torques at the ankle, knee, and hip joints were reduced by 30.9%, 41.8%, and 29.7%, respectively.

Finally, it should be noted that the Hybrid NMPC-DNN-PI control system has a very low computation time and, unlike standard NMPC, can operate in real time. The processing time per control step for the conditions studied in this research, run on a Ryzen 3 3200U, was approximately 0.04 ms, a negligible value, while for standard NMPC under the same conditions, it was around 60 ms. As a result, the computational cost was reduced by 99.93%, demonstrating that the control system is extremely fast while maintaining NMPC-level performance, making it suitable for replacing NMPC in various dynamic systems.

## 5. CONCLUSION

This study presented a new control system, Hybrid NMPC-DNN-PI. The deep neural network (DNN) approximated the nonlinear model predictive control (NMPC), significantly reducing computational cost, while the PI component assisted the neural network in the presence of disturbances and deviations from the training data. The controller was validated on a human-robot model during squat movement with three active joints. Due to the complexity of the dynamic model and the lack of robust nonlinear controllers for such exoskeleton robots in the literature, this setup was the most appropriate for validating the proposed control system.

The control objective was to compensate for human gravitational torques and help reduce joint torques. Under unknown conditions and external disturbances, Hybrid NMPC-DNN-PI outperformed NMPC-DNN (the NMPC approximated by a DNN) in reducing tracking errors. Moreover, the controller significantly decreased human joint torques compared to the case without the robot. For a particular squat movement trajectory in the dataset, corresponding to a human with a weight of 80 kg, height of 1.90 m, and a 1.75 s squat cycle, the RMS torques at the ankle, knee, and hip were reduced by 30.9%, 41.8%, and 29.7%, respectively. Finally, the computational cost of the proposed control system was 99.93% lower than that of NMPC, demonstrating the potential of the Hybrid NMPC-DNN-PI framework for real-time control. Future works could explore using other neural networks, such as CNN or LSTM, to approximate NMPC behavior or apply the proposed control system to other dynamic systems.

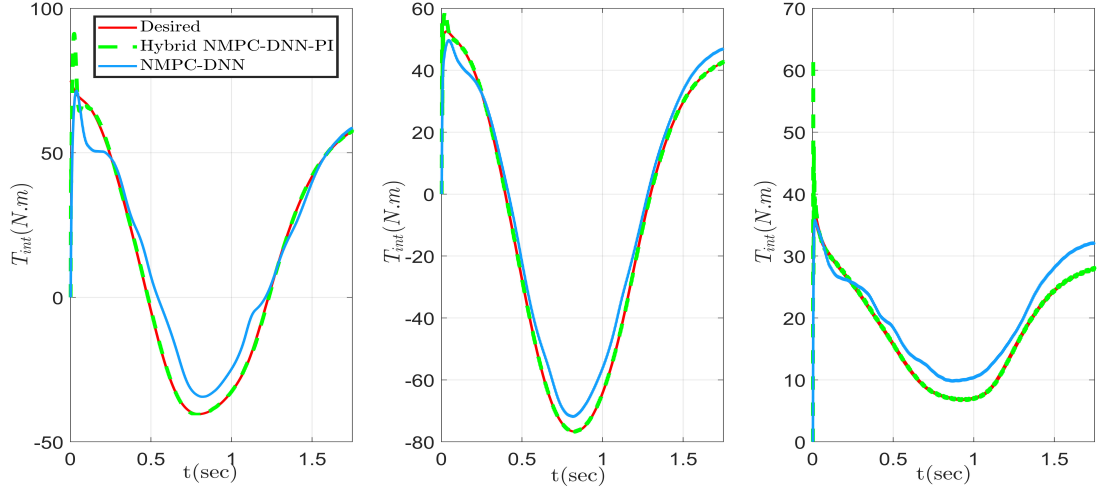

Fig. 5. Tracking of the desired interaction torques by the Hybrid NMPC-DNN-PI and NMPC-DNN control systems for the shank (left), thigh (middle), and upper body (right).

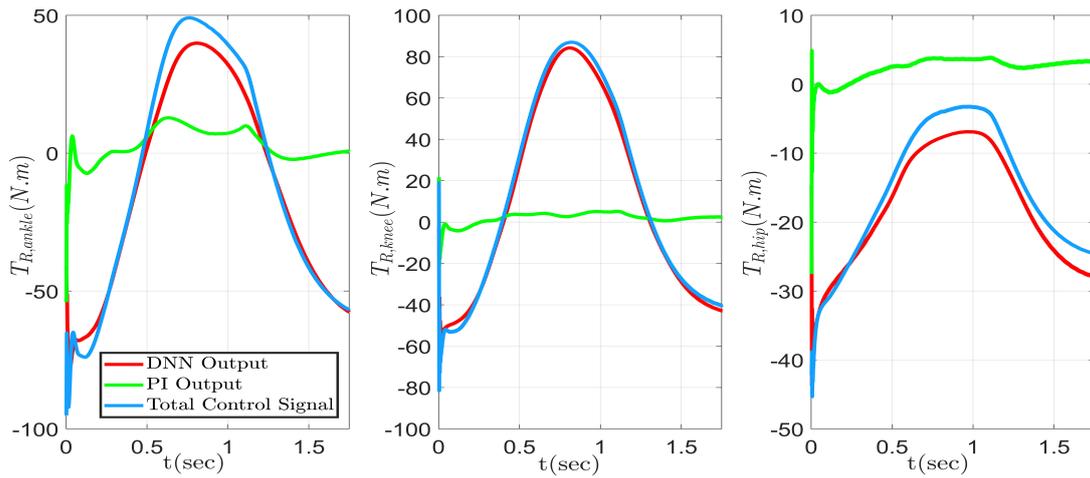

Fig. 6. Torques calculated by the proposed control system, showing contributions from the DNN and added PI term for the ankle (left), knee (middle), and hip (right).

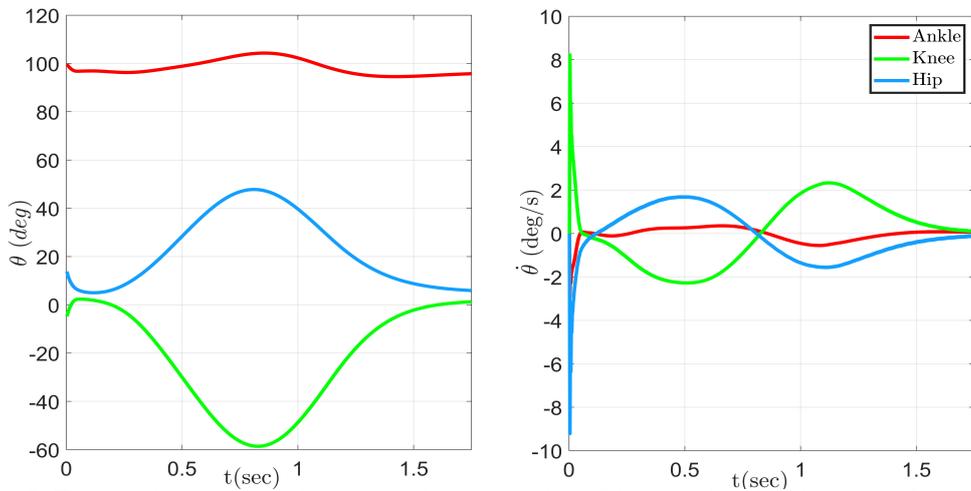

Fig. 7. The angular positions (left) and velocities (right) of the robot's ankle, knee, and hip joints.

# CONFLICT OF INTEREST

The authors declare no conflicts of interest exist. The authors declare that there is no competing financial interest or personal relationship that could have appeared to influence the work reported in this paper.

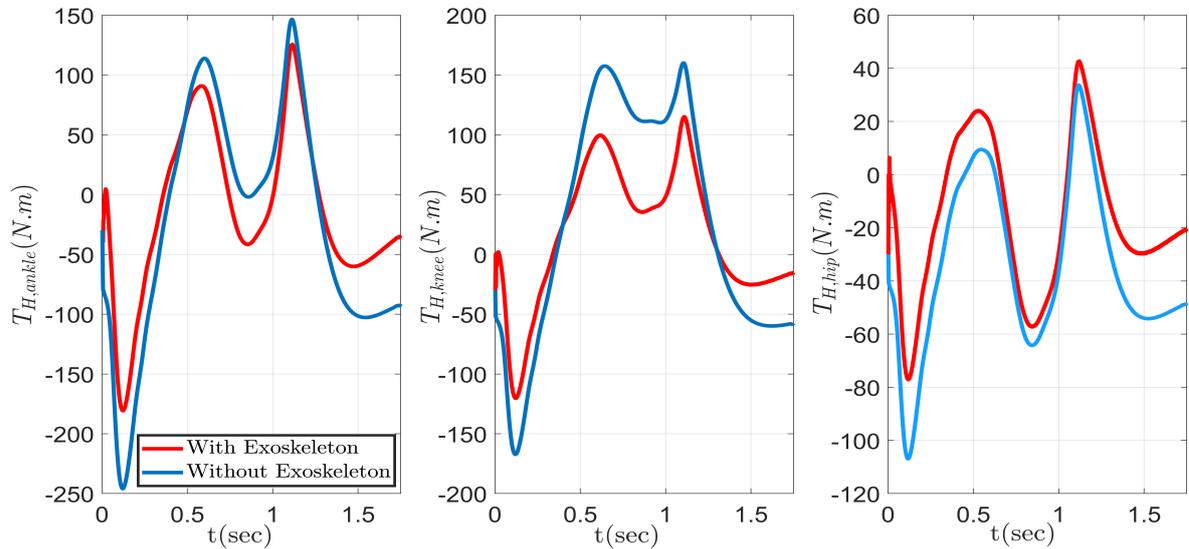

Fig. 8. Comparison of human joint torques during squat with and without the Hybrid NMPC-DNN-PI controlled exoskeleton for the ankle (left), knee (middle), and hip (right).

## AUTHOR CONTRIBUTION

Conceptualization: AA, Data curation: AA, Formal analysis: AA, Funding acquisition: GV, Investigation: AA, Methodology: AA, Project administration: GV, AA, Software: AA, Resources: AA, Supervision: GV, Validation: AA, Visualization: AA, Writing – original draft: AA, Writing – review & editing: AA.

## REFERENCES

[1] Michael G. Forbes, Rohit S. Patwardhan, Hamza Hamadah, R. Bhushan Gopaluni, "Model Predictive Control in Industry: Challenges and Opportunities, " IFAC-PapersOnLine, Volume 48, Issue 8, 2015, Pages 531-538, ISSN 2405-8963.

[2] Michael Nikolaou, "Model predictive controllers: A critical synthesis of theory and industrial needs", Advances in Chemical Engineering, Academic Press, 2001, Volume 26, Pages 131-204.

[3] Karlsson, H.; Hagentoft, C.E. "Application of model based predictive control for water-based floor heating in low energy residential buildings". Build. Environ. 2011, 46, 556–569.

[4] Yuan, S.; Perez, R. "Multiple-zone ventilation and temperature control of a single-duct VAV system using model predictive strategy". Energy Build. 2006, 38, 1248–1261.

[5] Behrooz, F.; Mariun, N.; Marhaban, M.H.; Mohd Radzi, M.A.; Ramli, A.R. "Review of Control Techniques for HVAC Systems—Nonlinearity Approaches Based on Fuzzy Cognitive Maps". Energies 2018, 11, 495.

[6] Behrooz, F.; Mariun, N.; Marhaban, M.H.; Mohd Radzi, M.A.; Ramli, A.R. "Review of Control Techniques for HVAC Systems—Nonlinearity Approaches Based on Fuzzy Cognitive Maps". Energies 2018, 11, 495.

[7] Marusak, P.M. "Numerically Efficient Fuzzy MPC Algorithm with Advanced Generation of Prediction-Application to a Chemical Reactor. " Algorithms 2020, 13, 143.

[8] L. Wang, E. H. F. van Asseldonk and H. van der Kooij, "Model predictive control-based gait pattern generation for wearable exoskeletons," 2011 IEEE International Conference on Rehabilitation Robotics, 2011, pp. 1-6.

[9] S. Subramanian, S. Lucia and S. Engell, "Economic multi-stage output nonlinear model predictive control", 2014 IEEE Conference on Control Applications (CCA), 2014, pp. 1837-1842.

[10] D. Q. Mayne, E. C. Kerrigan, E. J. van Wyk, and P. Falugi. "Tube based robust nonlinear model predictive control". International Journal of Robust and Nonlinear Control, 21(11):1341–1353, 2011.

[11] Ryuta Moriyasu, Sayaka Nojiri, Akio Matsunaga, Toshihiro Nakamura, Tomohiko Jimbo, "Diesel engine air path control based on neural approximation of nonlinear MPC", Control Engineering Practice, Volume 91, 2019, 104114, ISSN 0967-0661.

[12] Rajabi, Farshad, Behrooz Rezaie, and Zahra Rahmani. "A novel nonlinear model predictive control design based on a hybrid particle swarm optimization-sequential quadratic programming algorithm: Application to an evaporator system." Transactions of the Institute of Measurement and Control 38.1 (2016): 23-32.

[13] Aliyari, Alireza and Gholamreza Vossoughi. "Multi-stage robust nonlinear model predictive control of a lower-limb exoskeleton robot." arXiv preprint arXiv:2509.22120 (2025).

[14] Sergio Lucia, Benjamin Karg, "A deep learning-based approach to robust nonlinear model predictive control", IFAC-PapersOnLine, Volume 51, Issue 20, 2018, Pages 511-516, ISSN 2405-8963.

[15] A. Tagliabue, D. -K. Kim, M. Everett and J. P. How, "Demonstration-Efficient Guided Policy Search via Imitation of Robust Tube MPC," 2022 International Conference on Robotics and Automation (ICRA), Philadelphia, PA, USA, 2022, pp. 462-468.

[16] Kermanshah, Mehdi, et al. "Model Predictive Control for Magnetically-Actuated Cellbots." arXiv preprint arXiv:2406.02722 (2024).

[17] Mollahossein, Mojtaba, et al. "Attention-Based Convolutional Neural Network Model for Human Lower Limb Activity Recognition using sEMG." arXiv preprint arXiv:2506.06624 (2025).

[18] Shabani, Sharif, et al. "Development of a Stereo Vision-based UGV Guidance System For Bareroot Forest Nurseries." Smart Agricultural Technology (2025): 100990.

[19] J. E. Pratt, B. T. Krupp, C. J. Morse, and S. H. Collins, "The RoboKnee: an exoskeleton for enhancing strength and endurance during walking," in Robotics and Automation, 2004. Proceedings. ICRA'04. 2004 IEEE International Conference on, 2004, pp. 2430-2435.

[20] Wei, Wei, Shijia Zha, Yuxuan Xia, Jihua Gu, and Xichuan Lin. 2020. "A Hip Active Assisted Exoskeleton That Assists the Semi-Squat Lifting" Applied Sciences 10, no. 7: 2424.

[21] Fatai Sado, Hwa Jen Yap, Raja Ariffin Raja Ghazilla, Norhafizan Ahmad, "Design and control of a wearable lower-body exoskeleton for squatting and walking assistance in manual handling works", Mechatronics, Volume 63, 2019, 102272.

[22] Luo S, Androwis G, Adamovich S, Su H, Nunez E, Zhou X. "Reinforcement Learning and Control of a Lower Extremity Exoskeleton for Squat Assistance", Front Robot AI. 2021 Jul 19; 8:702845.

[23] Yang, Xiaoliang, et al. "A predictive power control strategy for DFIGs based on a wind energy converter system." Energies 10.8 (2017): 1098.

[24] Gill, P.E., Murray, W. and Saunders, M.A., 2005. SNOPT: An SQP algorithm for large-scale constrained optimization. SIAM review, 47(1), pp.99-131.

[25] Winter, D.A., Biomechanics and motor control of human movement. 2009: John Wiley & Sons.

[26] Vilches-Ponce, K., F. Lara, and M. Mora. "Deep neural networks for solving differential equations: a brief review." Journal of Physics: Conference Series. Vol. 3117. No. 1. IOP Publishing, 2025.

[27] Riedmiller, M & Braun, H. (1993). A Direct Adaptive Method for Faster Backpropagation Learning: "The RPROP Algorithm". IEEE INTERNATIONAL CONFERENCE ON NEURAL NETWORKS. Retrieved October 31, 2011.

[28] Sibi, P., Jones, S.A. and Siddarth, P., 2013. "Analysis of different activation functions using back propagation neural networks." Journal of Theoretical and Applied Information Technology, 47(3), pp.1264-1268.

[29] Weitz, Andrew, et al. "InfiniteForm: A synthetic, minimal bias dataset for fitness applications." arXiv preprint arXiv:2110.01330 (2021).Aliman N, Ramli R, Haris SM.